\documentclass[11pt,a4paper]{article}
\usepackage{authblk}
\usepackage[hyperref]{acl2018}
\usepackage{times}
\usepackage{latexsym}
\usepackage{latexsym}
\usepackage{amsmath}
\usepackage{mathrsfs}
\usepackage{xspace}
\usepackage{graphicx}
\usepackage{algorithm, algorithmicx, algpseudocode}
\usepackage{url}
\usepackage{color}
\usepackage{boxedminipage2e}
\usepackage{multirow}
\usepackage{color,soul}
\usepackage{tablefootnote}
\usepackage{makecell}
\usepackage{array}

\aclfinalcopy 

\algnewcommand{\LeftComment}[1]{\Statex \(\triangleright\) #1}

\title{ Triangular Architecture for Rare Language Translation}

\author[1,2\thanks{\ \ Contribution during internship at MSRA.}]{Shuo Ren}
\author[3]{Wenhu Chen}
\author[4]{Shujie Liu}
\author[4]{Mu Li}
\author[4]{Ming Zhou}
\author[1,2]{Shuai Ma}
\affil[1]{SKLSDE Lab, Beihang University, Beijing, China}
\affil[2]{Beijing Advanced Innovation Center for Big Data and Brain Computing, Beijing, China}
\affil[3]{University of California, Santa Barbara, CA, USA}
\affil[4]{Microsoft Research in Asia, Beijing, China}

\newcommand{\xtoy}{\ensuremath{p(y|x)}\xspace}
\newcommand{\ytox}{\ensuremath{p(x|y)}\xspace}
\newcommand{\xtoz}{\ensuremath{p(z|x)}\xspace}
\newcommand{\ztox}{\ensuremath{p(x|z)}\xspace}
\newcommand{\ytoz}{\ensuremath{p(z|y)}\xspace}
\newcommand{\ztoy}{\ensuremath{p(y|z)}\xspace}

\date{}
\hypersetup{draft} 
\begin{document}

\maketitle
\begin{abstract}
Neural Machine Translation (NMT) performs poor on the low-resource language pair $(X,Z)$, especially when $Z$ is a rare language. By introducing another rich language $Y$, we propose a novel triangular training architecture (TA-NMT) to leverage bilingual data $(Y,Z)$ (may be small) and $(X,Y)$ (can be rich) to improve the translation performance of low-resource pairs. In this triangular architecture, $Z$ is taken as the intermediate latent variable, and translation models of $Z$ are jointly optimized with a unified bidirectional EM algorithm under the goal of maximizing the translation likelihood of $(X,Y)$. Empirical results demonstrate that our method significantly improves the translation quality of rare languages on MultiUN and IWSLT2012 datasets, and achieves even better performance combining back-translation methods.
\end{abstract}

\section{Introduction}
\label{Introduction}
In recent years, Neural Machine Translation (NMT) \cite{kalchbrenner2013recurrent,sutskever2014sequence,bahdanau2014neural} has achieved remarkable performance on many translation tasks \citep{jean2015montreal,sennrich2016edinburgh,wu2016google, sennrich2017university}. Being an end-to-end architecture, an NMT system first encodes the input sentence into a sequence of real vectors, based on which the decoder generates the target sequence word by word with the attention mechanism \cite{bahdanau2014neural,luong2015effective}. During training, NMT systems are optimized to maximize the translation probability of a given language pair with the Maximum Likelihood Estimation (MLE) method, which requires large bilingual data to fit the large parameter space. Without adequate data, which is common especially when it comes to a rare language, NMT usually falls short on low-resource language pairs~\cite{zoph2016transfer}. 

In order to deal with the data sparsity problem for NMT, exploiting monolingual data \cite{sennrich2015improving,zhang2016exploiting,cheng2016semi,zhang2017joint, he2016dual} is the most common method. With monolingual data, the back-translation method \cite{sennrich2015improving} generates pseudo bilingual sentences with a target-to-source translation model to train the source-to-target one. By extending back-translation, source-to-target and target-to-source translation models can be jointly trained and boost each other \cite{cheng2016semi,zhang2017joint}. Similar to joint training \cite{cheng2016semi,zhang2017joint}, dual learning \cite{he2016dual} designs a reinforcement learning framework to better capitalize on monolingual data and jointly train two models.

Instead of leveraging monolingual data ($X$ or $Z$) to enrich the low-resource bilingual pair $(X,Z)$, in this paper, we are motivated to introduce another rich language $Y$, by which additionally acquired bilingual data $(Y,Z)$ and $(X,Y)$ can be exploited to improve the translation performance of $(X,Z)$. This requirement is easy to satisfy, especially when $Z$ is a rare language but $X$ is not. Under this scenario, $(X,Y)$ can be a rich-resource pair and provide much bilingual data, while $(Y,Z)$ would also be a low-resource pair mostly because $Z$ is rare. For example, in the dataset IWSLT2012, there are only 112.6K bilingual sentence pairs of English-Hebrew, since Hebrew is a rare language. If French is introduced as the third language, we can have another low-resource bilingual data of French-Hebrew (116.3K sentence pairs), and easily-acquired bilingual data of the rich-resource pair English-French.

\begin{figure}[!htb]
\centering
\includegraphics[width=0.5\linewidth]{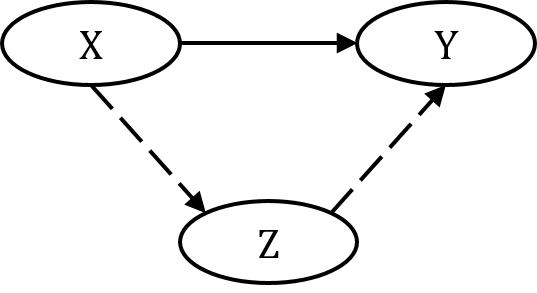}
\caption{Triangular architecture for rare language translation. The solid lines mean rich-resource and the dash lines mean low-resource. $X$, $Y$ and $Z$ are three different languages.}
\label{fig:Scenario}
\end{figure}

With the introduced rich language $Y$, in this paper, we propose a novel triangular architecture (TA-NMT) to exploit the additional bilingual data of $(Y,Z)$ and $(X,Y)$, in order to get better translation performance on the low-resource pair $(X,Z)$, as shown in \autoref{fig:Scenario}. In this architecture, $(Y,Z)$ is used for training another translation model to score the translation model of $(X,Z)$, while $(X,Y)$ is used to provide large bilingual data with favorable alignment information. 

Under the motivation to exploit the rich-resource pair $(X,Y)$, instead of modeling $X \Rightarrow Z$ directly, our method starts from modeling the translation task $X \Rightarrow Y$ while taking $Z$ as a latent variable. Then, we decompose $X \Rightarrow Y$ into two phases for training two translation models of low-resource pairs ($(X,Z)$ and $(Y,Z)$) respectively. The first translation model generates a sequence in the hidden space of $Z$ from $X$, based on which the second one generates the translation in $Y$. 
These two models can be optimized jointly with an Expectation Maximization (EM) framework with the goal of maximizing the translation probability \xtoy. In this framework, the two models can boost each other by generating pseudo bilingual data for model training with the weights scored from the other. By reversing the translation direction of $X \Rightarrow Y$, our method can be used to train another two translation models \ytoz and \ztox. Therefore, the four translation models (\xtoz, \ztox, \ytoz and \ztoy) of the rare language $Z$ can be optimized jointly with our proposed unified bidirectional EM algorithm. 

Experimental results on the MultiUN and IWSLT2012 datasets demonstrate that our method can achieve significant improvements for rare languages translation. By incorporating back-translation (a method leveraging more monolingual data) into our method, TA-NMT can achieve even further improvements.  

Our contributions are listed as follows:
\begin{itemize}
\item We propose a novel triangular training architecture (TA-NMT) to effectively tackle the data sparsity problem for rare languages in NMT with an EM framework.
\item Our method can exploit two additional bilingual datasets at both the model and data levels by introducing another rich language.
\item Our method is a unified bidirectional EM algorithm, in which four translation models on two low-resource pairs are trained jointly and boost each other.
\end{itemize}

\section{Method}

As shown in ~\autoref{fig:Scenario}, our method tries to leverage $(X,Y)$ (a rich-resource pair) and $(Y,Z)$ to improve the translation performance of low-resource pair $(X,Z)$, during which translation models of $(X,Z)$ and $(Y,Z)$ can be improved jointly.

Instead of directly modeling the translation probabilities of low-resource pairs, we model the rich-resource pair translation $X \Rightarrow Y$,  with the language $Z$ acting as a bridge to connect $X$ and $Y$. We decompose $X \Rightarrow Y$ into two phases for training two translation models. The first model \xtoz generates the latent translation in $Z$ from the input sentence in $X$, based on which the second one \ztoy generate the final translation in language $Y$. Following the standard EM procedure \cite{borman2004expectation} and Jensen's inequality, we derive the lower bound of $\xtoy$ over the whole training data $D$ as follows:
\begin{equation}
\begin{aligned}
& L(\Theta;D) \\
& = \sum_{(x, y) \in D}\log{\xtoy} \\
& = \sum_{(x, y) \in D}\log{\sum_{z}p(z|x)p(y|z)}\\
& = \sum_{(x, y) \in D}\log{\sum_{z}Q(z)\frac{\xtoz p(y|z)}{Q(z)}}\\
& \geq \sum_{(x, y) \in D}\sum_{z}Q(z)\log \frac{\xtoz p(y|z)}{Q(z)}\\
&\doteq\mathcal{L}(Q) \\
\end{aligned}\label{Jensen's}
\end{equation}
where $\Theta$ is the model parameters set of $\xtoz$ and $\ztoy$, and $Q(z)$ is an arbitrary posterior distribution of $z$. We denote the lower-bound in the last but one line as $\mathcal{L}(Q)$. Note that we use an approximation that $p(y|x,z)\approx\ztoy$ due to the semantic equivalence of parallel sentences $x$ and $y$.

In the following subsections, we will first propose our EM method  in \autoref{Original Model} based on the lower-bound derived above. Next, we will extend our method to two directions and give our unified bidirectional EM training in \autoref{bidirectional}. Then, in \autoref{Training_methods}, we will discuss more training details of our method and present our algorithm in the form of pseudo codes.

\subsection{EM Training}
\label{Original Model}
To maximize $L(\Theta;D)$, the EM algorithm can be leveraged to maximize its lower bound $\mathcal{L}(Q)$. In the E-step, we calculate the expectation of the variable $z$ using current estimate for the model, namely find the posterior distribution $Q(z)$. In the M-step, with the expectation $Q(z)$, we maximize the lower bound $\mathcal{L}(Q)$. Note that conditioned on the observed data and current model, the calculation of $Q(z)$ is intractable, so we choose $Q(z)=\xtoz$ approximately.

\textbf{M-step: }In the M-step, we maximize the lower bound $\mathcal{L}(Q)$ w.r.t model parameters given $Q(z)$. By substituting $Q(z)=\xtoz$ into $\mathcal{L}(Q)$, we can get the M-step as follows:
\begin{equation}
\begin{aligned}
\Theta_{y|z}
&=\mathop{\arg\max}_{\Theta_{y|z}}\mathcal{L}(Q) \\
&=\mathop{\arg\max}_{\Theta_{y|z}} \sum_{(x, y) \in D}\sum_{z}\xtoz\log{\ztoy}\\
&=\mathop{\arg\max}_{\Theta_{y|z}}\sum_{(x, y) \in D}E_{z\sim{\xtoz}}\log{\ztoy}
\end{aligned}\label{M-step1}
\end{equation}

\textbf{E-step: }The approximate choice of $Q(z)$ brings in a gap between $\mathcal{L}(Q)$ and $L(\Theta;D)$, which can be minimized in the E-step with Generalized EM method \cite{mclachlan2007algorithm}. 
According to \newcite{bishop2006pattern}, we can write this gap explicitly as follows:
\begin{equation}
\begin{aligned}
L(\Theta;D) - \mathcal{L}(Q) & = \sum_{z}Q(z)\log \frac{Q(z)}{p(z|y)}\\
& = KL(Q(z)||p(z|y))\\
& = KL(p(z|x)||p(z|y)) \\
\end{aligned}\label{decomposition}
\end{equation}
where $KL(\cdot)$ is the Kullback–Leibler divergence, and the approximation that $p(z|x,y)\approx{p(z|y)}$ is also used above. 

In the E-step, we minimize the gap between $\mathcal{L}(Q)$ and $L(\Theta;D)$ as follows:
\begin{equation}
\begin{aligned}
\Theta_{z|x}
&=\mathop{\arg\min}_{\Theta_{z|x}} KL(p(z|x)||p(z|y)) \\
\end{aligned}\label{E-step1}
\end{equation}

To sum it up, the E-step optimizes the model $\xtoz$ by minimizing the gap between $\mathcal{L}(Q)$ and $L(\Theta;D)$ to get a better lower bound $\mathcal{L}(Q)$. This lower bound is then maximized in the M-step to optimize the model $\ztoy$. Given the new model $\ztoy$, the E-step tries to optimize $\xtoz$ again to find a new lower bound, with which the M-step is re-performed. This iteration process continues until the models converge, which is guaranteed by the convergence of the EM algorithm.

\subsection{Unified Bidirectional Training}
\label{bidirectional}
The model $\ytoz$ is used as an approximation of $p(z|x,y)$ in the E-step optimization (\autoref{decomposition}). Due to the low resource property of the language pair $(Y,Z)$, $\ytoz$ cannot be well trained. To solve this problem, we can jointly optimize $\ztox$ and $\ytoz$ similarly by maximizing the reverse translation probability $\ytox$.

We now give our unified bidirectional generalized EM procedures as follows:

\begin{itemize}
\item Direction of $X \Rightarrow Y$

E: Optimize $\Theta_{z|x}$.
\begin{equation}
 \mathop{\arg\min}_{\Theta_{z|x}}KL(p(z|x)||p(z|y))\label{E-step1}
\end{equation}
M:  Optimize $\Theta_{y|z}$.
\begin{equation}
\mathop{\arg\max}_{\Theta_{y|z}}\sum_{(x, y) \in D}E_{z\sim{\xtoz}}\log{\ztoy}\label{M-step1}
\end{equation}

\item Direction of $Y \Rightarrow X$

E: Optimize $\Theta_{z|y}$.
\begin{equation}
\mathop{\arg\min}_{\Theta_{z|y}}KL(p(z|y)||p(z|x))\label{E-step2}
\end{equation}
M: Optimize $\Theta_{x|z}$.
\begin{equation}
\mathop{\arg\max}_{\Theta_{x|z}}\sum_{(x, y) \in D}E_{z\sim{\ytoz}}\log{\ztox}\label{M-step2}
\end{equation}
\end{itemize}

\begin{figure}[!htb]
\centering
\includegraphics[width=0.7\linewidth]{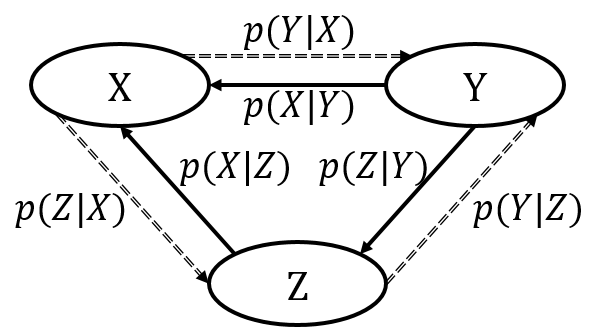}
\caption{Triangular Learning Architecture for Low-Resource NMT}
\label{fig:Model_Architecture}
\end{figure}

Based on the above derivation, the whole architecture of our method can be illustrated in ~\autoref{fig:Model_Architecture}, where the dash arrows denote the direction of $\xtoy$, in which $\xtoz$ and $\ztoy$ are trained jointly with the help of $\ytoz$, while the solid ones denote the direction of $\ytox$, in which $\ytoz$ and $\ztox$ are trained jointly with the help of $\xtoz$.

\subsection{Training Details}
\label{Training_methods}
A major difficulty in our unified bidirectional training is the exponential search space of the translation candidates, which could be addressed by either sampling \cite{shen2015minimum, cheng2016semi} or mode approximation \cite{kim2016sequence}. In our experiments, we leverage the sampling method and simply generate the top target sentence for approximation.

In order to perform gradient descend training, the parameter gradients for Equations ~\ref{E-step1} and ~\ref{E-step2} are formulated as follows:
\begin{equation}
\begin{aligned}
&\nabla_{\Theta_{z|x}}KL(\xtoz||\ytoz) \\
&=
E_{z\sim {\xtoz}}\log \frac{{\xtoz}}{{\ytoz}}\nabla _{\Theta_{z|x}}\log{\xtoz}\\
&\nabla_{\Theta_{z|y}}KL(\ytoz||\xtoz) \\
&=
E_{z\sim {\ytoz}}\log \frac{{\ytoz}}{{\xtoz}}\nabla _{\Theta_{z|y}}\log{\ytoz}
\end{aligned}
\label{gradient}
\end{equation}

Similar to reinforcement learning, models $\xtoz$ and $\ytoz$ are trained using samples generated by the models themselves. According to our observation, some samples are noisy and detrimental to the training process. One way to tackle this is to filter out the bad ones using some additional metrics (BLEU, etc.). Nevertheless, in our settings, BLEU scores cannot be calculated during training due to the absence of the golden targets ($z$ is generated based on $x$ or $y$ from the rich-resource pair $(x,y)$). Therefore we choose IBM model1 scores to weight the generated translation candidates, with the word translation probabilities calculated based on the given bilingual data (the low-resource pair $(x,z)$ or $(y,z)$). Additionally, to stabilize the training process, the pseudo samples generated by model $\xtoz$ or $\ytoz$ are mixed with true bilingual samples in the same mini-batch with the ratio of 1-1.
The whole training procedure is described in the following Algorithm ~\ref{alg:1}, where the 5th and 9th steps are generating pseudo data.

\begin{algorithm}[!htb]
    \renewcommand{\algorithmicrequire}{\textbf{Input:}}
    \renewcommand{\algorithmicensure}{\textbf{Output:}}
    \caption{Training low-resource translation models with the triangular architecture} 
    \label{alg:1}
    \begin{algorithmic}[1]
        \label{Algorithm_description}
        \Require Rich-resource bilingual data $(x,y)$; low-resource bilingual data $(x,z)$ and $(y,z)$
        \Ensure Parameters $\Theta_{z|x}$, $\Theta_{y|z}$, $\Theta_{z|y}$ and $\Theta_{x|z}$
        \State Pre-train $\xtoz$, $\ytoz$, $\ztox$, $\ztoy$     
        \While {not convergence}
        \State Sample $\mathop{(x, y)}, \mathop{(x^*,z^*)}, \mathop{(y^*,z^*)} \in D$
        \State \Comment{$X \Rightarrow Y$: Optimize $\Theta_{z|x}$ and $\Theta_{y|z}$}
        \State Generate $\mathop{z'}$ from $\mathop{p(z'|x)}$ and build the training batches $B_1 = (x, z')\cup(x^*, z^*)$, $B_2 = (y, z')\cup(y^*,z^*)$  
        \State E-step: update $\Theta_{z|x}$ with $B_1$ (\autoref{E-step1}) 
        \State M-step: update $\Theta_{y|z}$ with $B_2$ (\autoref{M-step1})
       \State \Comment{$Y \Rightarrow X$: Optimize $\Theta_{z|y}$ and $\Theta_{x|z}$}
        \State Generate $\mathop{z'}$ from $\mathop{p(z'|y)}$ and build the training batches $B_3 = (y, z')\cup(y^*, z^*)$, $B_4 = (x, z')\cup(x^*,z^*)$
        \State E-step: update $\Theta_{z|y}$  with $B_3$ (\autoref{E-step2})
        \State M-step: update $\Theta_{x|z}$  with $B_4$ (\autoref{M-step2})
        \EndWhile
        \State \textbf{return}  $\Theta_{z|x}$, $\Theta_{y|z}$, $\Theta_{z|y}$ and $\Theta_{x|z}$
    \end{algorithmic}
\end{algorithm}

\section{Experiments}
\subsection{Datasets}
In order to verify our method, we conduct experiments on two multilingual datasets. The one is MultiUN~\cite{MultiUN}, which is a collection of translated documents from the United Nations, and the other is IWSLT2012~\cite{cettolo2012wit3}, which is a set of multilingual transcriptions of TED talks. As is mentioned in ~\autoref{Introduction}, our method is compatible with methods exploiting monolingual data. So we also find some extra monolingual data of rare languages in both datasets and conduct experiments incorporating back-translation into our method.

\textbf{MultiUN:} English-French (EN-FR) bilingual data are used as the rich-resource pair $(X,Y)$. Arabic (AR) and Spanish (ES) are used as two simulated rare languages $Z$. We randomly choose subsets of bilingual data of $(X,Z)$ and $(Y,Z)$ in the original dataset to simulate low-resource situations, and make sure there is no overlap in $Z$ between chosen data of $(X,Z)$ and $(Y,Z)$.

\textbf{IWSLT2012\footnote{https://wit3.fbk.eu/mt.php?release=2012-02-plain}:}  English-French is used as the rich-resource pair $(X,Y)$, and two rare languages $Z$ are Hebrew (HE) and Romanian (RO) in our choice. Note that in this dataset, low-resource pairs $(X,Z)$ and $(Y,Z)$ are severely overlapped in $Z$. In addition, English-French bilingual data from WMT2014 dataset are also used to enrich the rich-resource pair. We also use additional English-Romanian bilingual data from Europarlv7 dataset ~\cite{koehn2005europarl}. The monolingual data of $Z$ (HE and RO) are taken from the web\footnote{https://github.com/ajinkyakulkarni14/TED-Multilingual-Parallel-Corpus}. 

In both datasets, all sentences are filtered within the length of 5 to 50 after tokenization. Both the validation and the test sets are 2,000 parallel sentences sampled from the bilingual data, with the left as training data. The size of training data of all language pairs are shown in ~\autoref{tab:dataset}. 
\begin{table}
\centering
\resizebox{7.6cm}{!}{
\begin{tabular}{c|c|c||c|c}
\multirow{2}{*}{Pair} 
    & \multicolumn{2}{c||}{MultiUN} & \multicolumn{2}{c}{IWSLT2012} \\
    \cline{2-5}
       & Lang & Size & Lang & Size \\
\hline
\hline
$(X,Y)$ & EN-FR & 9.9 M & EN-FR \tablefootnote{together with WMT2014} & 7.9 M \\
\hline
$(X,Z)$ & EN-AR & 116 K & EN-HE & 112.6 K \\
$(Y,Z)$ & FR-AR & 116 K & FR-HE & 116.3 K \\
mono $Z$ & AR & 3 M & HE & 512.5 K \\
\hline
$(X,Z)$ & EN-ES & 116 K & EN-RO \tablefootnote{together with Europarlv7} & 467.3 K \\
$(Y,Z)$ & FR-ES & 116 K & FR-RO & 111.6 K \\
mono $Z$ & ES & 3 M & RO & 885.0 K \\
\end{tabular}}
\caption{\label{tab:dataset} training data size of each language pair.}
\end{table}

\subsection{Baselines}
\label{baselines}

We compare our method with four baseline systems. The first baseline is the ~\textbf{RNNSearch} model \cite{bahdanau2014neural}, which is a sequence-to-sequence model with attention mechanism trained with given small-scale bilingual data. The trained translation models are also used as pre-trained models for our subsequent training processes.

The second baseline is ~\textbf{PBSMT} \cite{koehn2003statistical}, which is a phrase-based statistical machine translation system. PBSMT is known to perform well on low-resource language pairs, so we want to compare it with our proposed method. And we use the public available implementation of Moses\footnote{http://www.statmt.org/moses/} for training and test in our experiments.

\begin{figure}[!htb]
\centering
\includegraphics[width=1.0\linewidth]{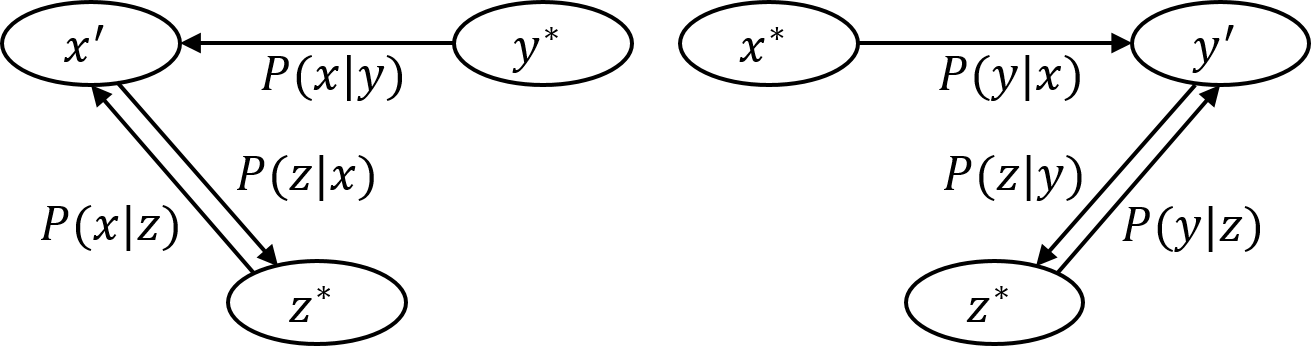}
\caption{A teacher-student alike method for low-resource translation. For training \xtoz and \ztox, we mix the true pair $(y^*,z^*) \in D$ with the pseudo pair $\mathop{(x',z^*)}$ generated by teacher model $p\mathop{(x'|y^*)}$ in the same mini-batch. The training procedure of $\ytoz$ and $\ztoy$ is similar.}
\label{fig:T-S}
\end{figure}

The third baseline is a teacher-student alike method \cite{chen2017teacher}. For the sake of brevity, we will denote it as ~\textbf{T-S}. The process is illustrated in ~\autoref{fig:T-S}. We treat this method as a second baseline because it can also be regarded as a method exploiting $(Y,Z)$ and $(X,Y)$ to improve the translation of $(X,Z)$ if we regard $(X,Z)$ as the zero-resource pair and $\ytox$ as the teacher model when training $\xtoz$ and $\ztox$. 

The fourth baseline is back-translation ~\cite{sennrich2015improving}. We will denote it as \textbf{BackTrans}. More concretely, to train the model \xtoz, we use extra monolingual $Z$ described in ~\autoref{tab:dataset} to do back-translation; to train the model \ztox, we use monolingual $X$ taken from $(X,Y)$. Procedures for training \ytoz and \ztoy are similar. This method use extra monolingual data of $Z$ compared with our TA-NMT method. But we can incorporate it into our method.

\subsection{Overall Results} 
\label{Overall_Results}
Experimental results on both datasets are shown in ~\autoref{tab:Results_MultiUN} and ~\ref{tab:Results_IWSLT} respectively, in which \textbf{RNNSearch}, \textbf{PBSMT}, \textbf{T-S} and \textbf{BackTrans} are four baselines. \textbf{TA-NMT} is our proposed method, and \textbf{TA-NMT(GI)} is our method incorporating back-translation as good initialization. For the purpose of clarity and a fair comparison, we list the resources that different methods exploit in \autoref{tab:resource}. 

\begin{table}
\centering
\resizebox{7.6cm}{!}{
\begin{tabular}{l|l}
Method & Resources \\
\hline\hline
PBSMT & $(X,Z)$, $(Y,Z)$ \\
\hline
RNNSearch & $(X,Z)$, $(Y,Z)$ \\
\hline
T-S & $(X,Z)$, $(Y,Z)$, $(X,Y)$\\
\hline
BackTrans & $(X,Z)$, $(Y,Z)$, $(X,Y)$, mono $Z$ \\
\hline
TA-NMT & $(X,Z)$, $(Y,Z)$, $(X,Y)$ \\
\hline
TA-NMT(GI) & $(X,Z)$, $(Y,Z)$, $(X,Y)$, mono $Z$
\end{tabular}}
\caption{\label{tab:resource} Resources that different methods use}
\end{table}

\renewcommand\arraystretch{1.2}
\begin{table*}[t!]
\small
\begin{center}
\begin{tabular}{l|c|c|c|c|c||c|c|c|c|c}
\multirow{2}*{Method}
 & EN2AR & AR2EN & FR2AR & AR2FR & \multirow{2}*{Ave} & EN2ES & ES2EN & FR2ES & ES2FR & \multirow{2}*{Ave}\\
 & (X$\Rightarrow$Z) & (Z$\Rightarrow$X) & (Y$\Rightarrow$Z) & (Z$\Rightarrow$Y) &  & (X$\Rightarrow$Z) & (Z$\Rightarrow$X) & (Y$\Rightarrow$Z) & (Z$\Rightarrow$Y) &  \\
\hline
\hline
RNNSearch & 18.03 & 31.40 & 13.42 & 22.04 & 21.22 & 38.77 & 36.51 & 32.92 & 33.05 & 35.31\\
PBSMT & 19.44 & 30.81 & 15.27 & 23.65 & 22.29 & 38.47 & 36.64 & 34.99 & 33.98 & 36.02\\
T-S & 19.02 & 32.47 & 14.59 & 23.53 & 22.40 & 39.75 & 38.02 & 33.67 & 34.04 & 36.57\\
BackTrans & 22.19 & 32.02 & 15.85 & 23.57 & 23.73 & 42.27 & 38.42 & 35.81 & 34.25 & 37.76 \\
\hline
TA-NMT & 20.59 & 33.22 & 14.64 & 24.45 & 23.23 & 40.85 & 39.06 & 34.52 & 34.39 & 37.21\\
TA-NMT(GI) & \bf23.16 & \bf33.64 & \bf16.50 & \bf25.07 & \bf24.59 & \bf42.63 & \bf39.53 & \bf35.87 & \bf35.21 & \bf38.31 \\
\end{tabular}
\end{center}
\caption{\label{tab:Results_MultiUN}Test BLEU on MultiUN Dataset.}
\end{table*}

\begin{table*}[t!]
\small
\begin{center}
\begin{tabular}{l|c|c|c|c|c||c|c|c|c|c}
\multirow{2}*{Method}
 & EN2HE & HE2EN & FR2HE & HE2FR & \multirow{2}*{Ave} & EN2RO & RO2EN & FR2RO & RO2FR & \multirow{2}*{Ave} \\
 & (X$\Rightarrow$Z) & (Z$\Rightarrow$X) & (Y$\Rightarrow$Z) & (Z$\Rightarrow$Y) & &(X$\Rightarrow$Z) & (Z$\Rightarrow$X) & (Y$\Rightarrow$Z) & (Z$\Rightarrow$Y) & \\
\hline
\hline
RNNSearch & 17.94 & 28.32 & 11.86 & 21.67 & 19.95 & 31.44 & 40.63 & 17.34 & 25.20 & 28.65\\
PBSMT & 17.39 & 28.05 & 12.77 & 21.87 & 20.02 & 31.51 & 39.98 & 18.13 & 25.47 & 28.77\\
T-S & 17.97 & 28.42 & 12.04 & 21.99 & 20.11 & 31.80 & 40.86 & 17.94 & 25.69 & 29.07\\
BackTrans & 18.69 & 28.55 & 12.31 & 21.63 & 20.20 & 32.18 & 41.03 & 18.19 & 25.30 & 29.18 \\
\hline
TA-NMT & 19.19 & 29.28 & 12.76 & 22.62 & 20.96 & 33.65 & 41.93 & 18.53 & 26.35 & 30.12\\
TA-NMT(GI) & \bf19.90 & \bf29.94 & \bf13.54 & \bf23.25 &\bf 21.66 & \bf34.41 & \bf42.61 & \bf19.30 & \bf26.53  & \bf30.71 \\
\end{tabular}
\end{center}
\caption{\label{tab:Results_IWSLT} Test BLEU on IWSLT Dataset. }
\end{table*}

From ~\autoref{tab:Results_MultiUN} on MultiUN, the performance of RNNSearch is relatively poor. As is expected, PBSMT performs better than RNNSearch on low-resource pairs by the average of 1.78 BLEU. The T-S method which can doubling the training data for both $(X,Z)$ and $(Y,Z)$ by generating pseudo data from each other, leads up to 1.1 BLEU points improvement on average over RNNSearch. Compared with T-S, our method gains a further improvement of about 0.9 BLEU on average, because our method can better leverage the rich-resource pair $(X,Y)$. With extra large monolingual $Z$ introduced, BackTrans can improve the performance of $\xtoz$ and $\ytoz$ significantly compared with all the methods without monolingual $Z$. However TA-NMT is comparable with or even better than BackTrans for \ztox and \ztoy because both of the methods leverage resources from rich-resource pair $(X,Y)$, but BackTrans does not use the alignment information it provides. Moreover, with back-translation as good initialization, further improvement is achieved by TA-NMT(GI) of about 0.7 BLEU on average over BackTrans.

In ~\autoref{tab:Results_IWSLT}, we can draw the similar conclusion. However, different from MultiUN, in the EN-FR-HE group of IWSLT, $(X,Z)$ and $(Y,Z)$ are severely overlapped in $Z$. Therefore, T-S cannot improve the performance obviously (only about 0.2 BLEU) on RNNSearch because it fails to essentially double training data via the teacher model. As for EN-FR-RO, with the additionally introduced EN-RO data from Europarlv7, which has no overlap in RO with FR-RO, T-S can improve the average performance more than the EN-FR-HE group. TA-NMT outperforms T-S by 0.93 BLEU on average. Note that even though BackTrans uses extra monolingual $Z$, the improvements are not so obvious as the former dataset, the reason for which we will delve into in the next subsection. Again, with back-translation as good initialization, TA-NMT(GI) can get the best result.

Note that BLEU scores of TA-NMT are lower than BackTrans in the directions of X$\Rightarrow$Z and Y$\Rightarrow$Z. The reason is that the resources used by these two methods are different, as shown in Table 2. To do back translation in two directions (e.g., X$\Rightarrow$Z and Z$\Rightarrow$X), we need monolingual data from both sides (e.g., X and Z), however, in TA-NMT, the monolingual data of Z is not necessary. Therefore, in the translation of X$\Rightarrow$Z or Y$\Rightarrow$Z, BackTrans uses additional monolingual data of Z while TA-NMT does not, that is why BackTrans outperforms TA-NMT in these directions. Our method can leverage back translation as a good initialization, aka TA-NMT(GI) , and outperforms BackTrans on all translation directions.

The average test BLEU scores of different methods in each data group (EN-FR-AR, EN-FR-ES, EN-FR-HE, and EN-FR-RO) are listed in the column \textbf{Ave} of the tables for clear comparison.


\subsection{The Effect of Extra Monolingual Data}
\label{Monolingual_effect}

\begin{figure}[!htb]
\centering
\includegraphics[width=0.90\linewidth]{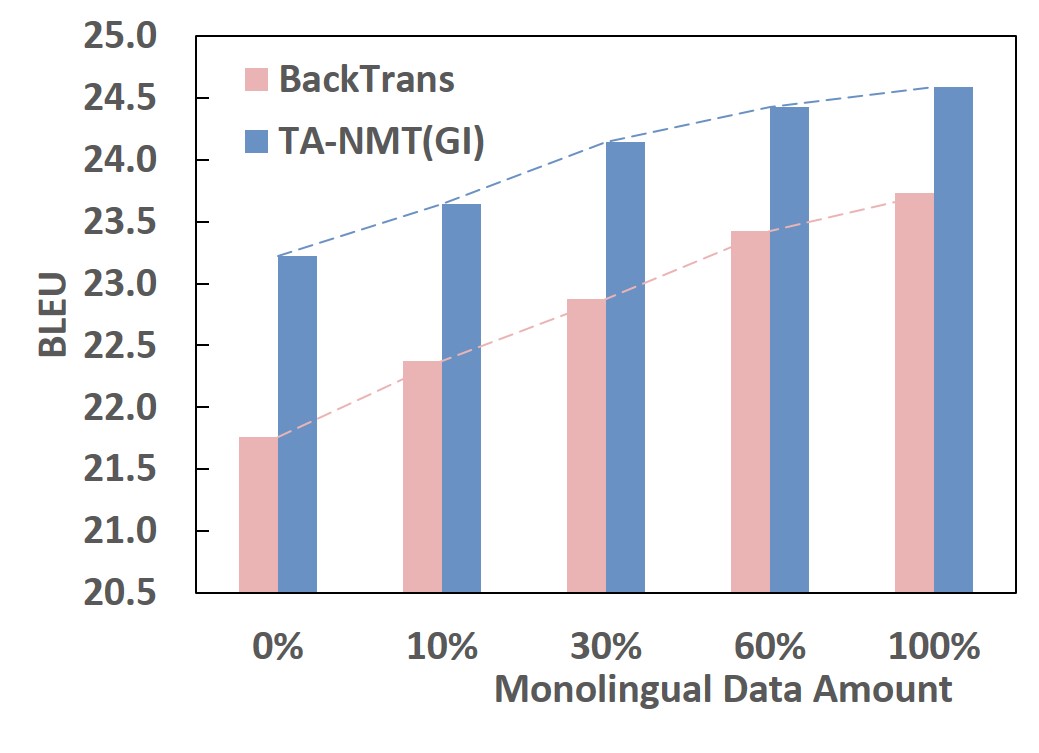}
\caption{Test BLEU of the EN-FR-AR group performed by BackTrans and TA-NMT(GI) with different amount of monolingual Arabic data.}
\label{fig:Monolingual}
\end{figure}

Comparing the results of BackTrans and TA-NMT(GI) on both datasets, we notice the improvements of both methods on IWSLT are not as significant as MultiUN. We speculate the reason is the relatively less amount of monolingual $Z$ we use in the experiments on IWSLT as shown in \autoref{tab:dataset}. So we conduct the following experiment to verify the conjecture by changing the scale of monolingual Arabic data in the MultiUN dataset, of which the data utilization rates are set to 0\%, 10\%, 30\%, 60\% and 100\% respectively. Then we compare the performance of BackTrans and TA-NMT(GI) in the EN-FR-AR group. As \autoref{fig:Monolingual} shows, the amount of monolingual $Z$ actually has a big effect on the results, which can also verify our conjecture above upon the less significant improvement of BackTrans and TA-NMT(GI) on IWSLT. In addition, even with poor "good-initialization", TA-NMT(GI) still get the best results. 

\subsection{EM Training Curves}
\begin{figure}[!htb]
\centering
\includegraphics[width=1.0\linewidth]{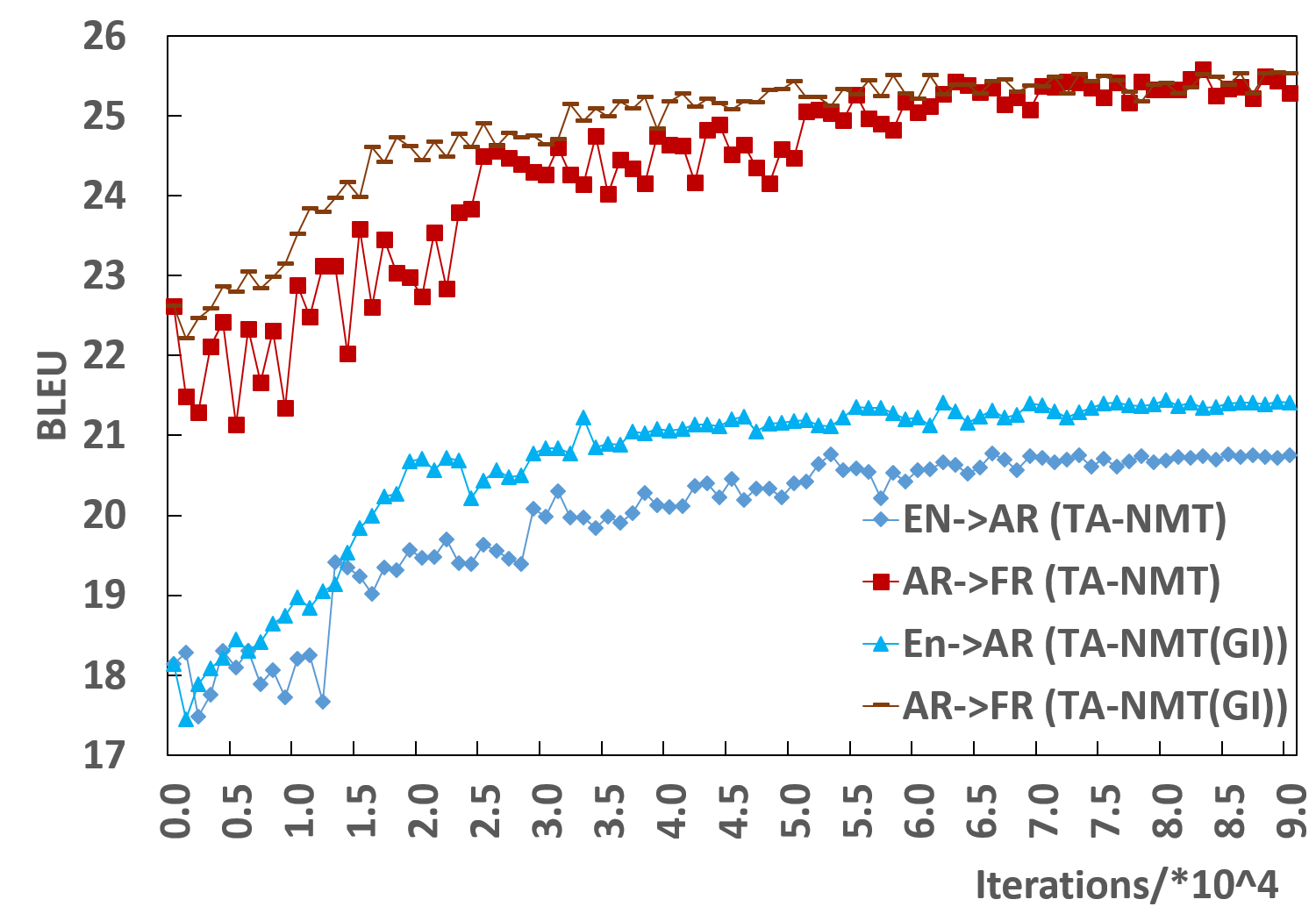}
\includegraphics[width=1.0\linewidth]{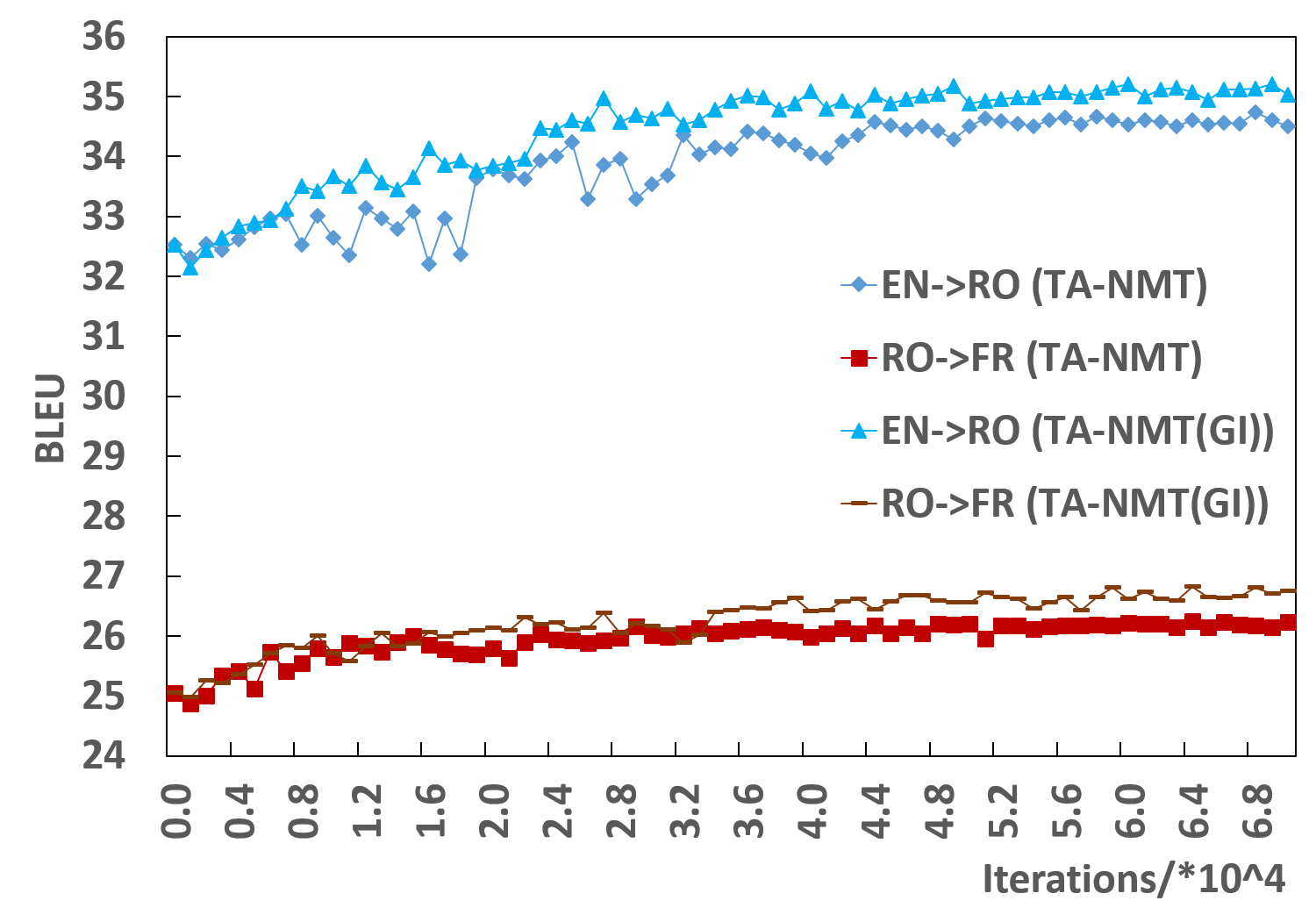}
\caption{BLEU curves on validation sets during the training processes of TA-NMT and TA-NMT(GI). (Top: EN2AR (the E-step) and AR2FR (the M-step); Bottom: EN2RO (the E-step) and RO2FR (the M-step))}
\label{fig:Valid_Curve}
\end{figure}

To better illustrate the behavior of our method, we print the training curves in both the M-steps and E-steps of TA-NMT and TA-NMT(GI) in ~\autoref{fig:Valid_Curve} above. The chosen models printed in this figure are EN2AR and AR2FR on MultiUN, and EN2RO and RO2FR on IWLST. 

From ~\autoref{fig:Valid_Curve}, we can see that the two low-resource translation models are improved nearly simultaneously along with the training process, which verifies our point that two weak models could boost each other in our EM framework. Notice that at the early stage, the performance of all models stagnates for several iterations, especially of TA-NMT. The reason could be that the pseudo bilingual data and the true training data are heterogeneous, and it may take some time for the models to adapt to a new distribution which both models agree. Compared with TA-NMT, TA-NMT(GI) are more stable, because the models may have adapted to a mixed distribution of heterogeneous data in the preceding back-translation phase.

\subsection{Reinforcement Learning Mechanism in Our Method}

\renewcommand\arraystretch{1.5}
\begin{table*}[t!]
\small
\begin{center}
\begin{tabular}{c|p{12cm}}
\hline  
Source & in concluding , poverty eradication requires {\color{blue}political} will and commitment .\\
\hline
\multirow{2}*{Output} 
    & en \underline{(0.66)} conclusi\'{o}n \underline{(0.80)} , \underline{(0.14)} la \underline{(0.00)} erradicaci\'{o}n \underline{(1.00)} de \underline{(0.40)} la \underline{(0.00)} pobreza \\ 
    & \underline{(0.90)} requiere \underline{(0.10)} voluntad \underline{(1.00)} y \underline{(0.46)} compromiso  \underline{(0.90)} {\color{blue}pol\'{i}ticas \underline{(-0.01)}} . (1.00) \\
\hline
Reference & \makecell[l]{en conclusi\'{o}n , la erradicaci\'{o}n de la pobreza necesita la voluntad y
compromiso {\color{blue}pol\'{i}ticos} .}\\       
\hline
\hline 
Source & \makecell[l]{{\color{blue}visit} us and get to know and {\color{blue}love berlin} !}\\
\hline
Output & \makecell[l]{{\color{blue}visita \underline{(0.00)}} y \underline{(0.05)} se \underline{(0.00)} a \underline{(0.17)} saber \underline{(0.00)} y \underline{(0.04)} {\color{blue}a \underline{(0.01)}}  
berl\'{i}n \underline{(0.00)} ! \underline{(0.00)}} \\
\hline
Reference & \makecell[l]{{\color{blue}vis\'{i}tanos} y llegar a saber y {\color{blue}amar a berl\'{i}n} .}\\   
\hline
\hline 
\multirow{2}*{Source}
	& legislation also provides an important means of recognizing economic , social and cultural \\ 
	& rights at the domestic level .\\
\hline
\multirow{3}*{Output}
    & la \underline{(1.00)} legislaci\'{o}n \underline{(0.34)} también \underline{(1.00)} constituye \underline{(0.60)} un \underline{(1.00)} medio \underline{(0.22)} importante \\
    & \underline{(0.74)} de \underline{(0.63)} reconocer \underline{(0.21)} los \underline{(0.01)} derechos \underline{(0.01)} 
económicos \underline{(0.03)} , \underline{(0.01)} sociales \\
     & \underline{(0.02)} y \underline{(0.01)} culturales \underline{(1.00)} a \underline{(0.00)} 
nivel \underline{(0.40)} nacional \underline{(1.00)} . \underline{(0.03)} \\
\hline
\multirow{2}*{Reference}
    & la legislaci\'{o}n tambi\'{e}n constituye un medio importante de reconocer los derechos econ\'{o}micos ,\\ 
    &iales y culturales a nivel nacional .\\ 
\hline
\end{tabular}
\end{center}
\caption{\label{tab:Instance_Analysis} English to Spanish translation sampled in the E-step as well as its time-step rewards. }
\end{table*}

As shown in \autoref{gradient}, the E-step actually works as a reinforcement learning (RL) mechanism. Models \xtoz and \ytoz generate samples by themselves and receive rewards to update their parameters. Note that the “reward” here is described by the log terms in \autoref{gradient}, which is derived from our EM algorithm rather than defined artificially. In ~\autoref{tab:Instance_Analysis}, we do a case study of the EN2ES translation sampled by \xtoz as well as its time-step rewards during the E-step. 

In the first case, the best translation of "political" is "pol\'{i}ticos". When the model \xtoz generates an inaccurate one "pol\'{i}ticas", it receives a negative reward (-0.01), with which the model parameters will be updated accordingly. In the second case, the output misses important words and is not fluent. Rewards received by the model \xtoz are zero for nearly all tokens in the output, leading to an invalid updating. In the last case, the output sentence is identical to the human reference. The rewards received are nearly all positive and meaningful, thus the RL rule will update the parameters to encourage this translation candidate.

\section{Related Work}
NMT systems, relying heavily on the availability of large bilingual data, result in poor translation quality for low-resource pairs \cite{zoph2016transfer}. This low-resource phenomenon has been observed in much preceding work. A very common approach is exploiting monolingual data of both source and target languages \cite{sennrich2015improving,zhang2016exploiting,cheng2016semi,zhang2017joint, he2016dual}.

As a kind of data augmentation technique, exploiting monolingual data can enrich the training data for low-resource pairs. \newcite{sennrich2015improving} propose back-translation, exploits the monolingual data of the target side, which is then used to generate pseudo bilingual data via an additional target-to-source translation model. Different from back-translation, \newcite{zhang2016exploiting} propose two approaches to use source-side monolingual data, of which the first is employing a self-learning algorithm to generate pseudo data, while the second is using two NMT models to predict the translation and to reorder the source-side monolingual sentences. As an extension to these two methods, \newcite{cheng2016semi} and \newcite{zhang2017joint} combine two translation directions and propose a training framework to jointly optimize the source-to-target and target-to-source translation models. Similar to joint training, \newcite{he2016dual} propose a dual learning framework with a reinforcement learning mechanism to better leverage monolingual data and make two translation models promote each other. All of these methods are concentrated on exploiting either the monolingual data of the source and target language or both of them. 

Our method takes a different angle but is compatible with existing approaches, we propose a novel triangular architecture to leverage two additional language pairs by introducing a third rich language. By combining our method with existing approaches such as back-translation, we can make a further improvement.

Another approach for tackling the low-resource translation problem is multilingual neural machine translation \cite{firat2016multi}, where different encoders and decoders for all languages with a shared attention mechanism are trained. This method tends to exploit the network architecture to relate low-resource pairs. Our method is different from it, which is more like a training method rather than network modification.

\section{Conclusion}
In this paper, we propose a triangular architecture (TA-NMT) to effectively tackle the problem of low-resource pairs translation with a unified bidirectional EM framework. By introducing another rich language, our method can better exploit the additional language pairs to enrich the original low-resource pair. Compared with the RNNSearch \cite{bahdanau2014neural}, a teacher-student alike method \cite{chen2017teacher} and the back-translation \cite{sennrich2015improving} on the same data level, our method achieves significant improvement on the MutiUN and IWSLT2012 datasets. Note that our method can be combined with methods exploiting monolingual data for NMT low-resource problem such as back-translation and make further improvements. 

In the future, we may extend our architecture to other scenarios, such as totally unsupervised training with no bilingual data for the rare language. 

\section*{Acknowledgments}
We thank Zhirui Zhang and Shuangzhi Wu for useful discussions. This work is supported in part by NSFC {\small U1636210}, 973 Program {\small 2014CB340300}, and NSFC {\small 61421003}.

\bibliography{acl2018}
\bibliographystyle{acl_natbib}

\appendix
\section{Implementation details}
\label{implementation}
All the NMT systems we used are implemented as the classic attention-based encoder-decoder framework with a bidirectional RNN encoder \cite{bahdanau2014neural}. The embedding size of both source and target words is 256, and hidden units of both encoder and decoder are 512-dimensional GRU cells for the MultiUN dataset and 256-dimensional for the IWSLT dataset. The vocabulary size is limited in 50K for each language in the MultiUN dataset and 30K in the IWSLT2012 dataset, with the out-of-vocabulary (OOV) words mapped to a special token $\langle unk \rangle$. The parameters are randomly initialized with sampling from the Gaussian distribution $N(0, 0.01)$.

We use mini-batch of size 64 with AdaDelta optimizer \cite{zeiler2012adadelta} for training . The learning rate in pre-training is set to 1.0 (the gradients are normalized), while in subsequent training stages it is set to 0.5. In the pre-training stage, we randomly shuffle the given data and train models for 20 to 30 epochs until converging.  In the test time, the beam search method is used for decoding and the beam size is set to 8.
\end{document}